\documentclass[a4paper]{article}
\usepackage{INTERSPEECH2020}

\usepackage{graphicx}
\usepackage{subfigure}
\usepackage{threeparttable}
\usepackage{tablefootnote}
\usepackage{url}

\graphicspath{{Fig/}}

\newcommand {\nvsum}{NV${_{sum}}\thinspace$}

\newcommand {\nvsumen}{NV${_{sum}^{(en)}}\thinspace $}

\newcommand {\nvsumde}{NV${_{sum}^{(de)}}\thinspace $}

\newcommand {\nvsumy}{NV${_{sum}^{(y)}}\thinspace $}

\newcommand {\nvori}{NV$_{:}$ \thinspace }

\newcommand {\nvsumconen}{NV${_{\star}^{(en)}}\thinspace $}
\newcommand {\nvsumconde}{NV${_{\star}^{(de)}}\thinspace $}
\newcommand {\nvsumcony}{NV${_{\star}^{(y)}}\thinspace $}
\newcommand {\nvsumcon}{NV${_{\star}}\thinspace $}

\newcommand {\cnvsum}{cNV${_{sum}}\thinspace $}
\newcommand {\cnvcon}{cNV${_{con}}\thinspace $}
\newcommand {\cnvsumen}{cNV${_{sum}^{(en)}}\thinspace $}

\newcommand {\cnvsumde}{cNV${_{sum}^{(de)}}\thinspace $}

\newcommand {\cnvsumy}{cNV${_{sum}^{(y)}}\thinspace $}
\newcommand {\cnvcony}{cNV${_{con}^{(y)}}\thinspace $}

\newcommand {\cnvsumcon}{cNV${_{\star}}\thinspace $}

\newcommand {\cnvsumconen}{cNV${_{\star}^{(en)}}\thinspace $}
\newcommand {\cnvsumconde}{cNV${_{\star}^{(de)}}\thinspace $}
\newcommand {\cnvsumcony}{cNV${_{\star}^{(y)}}\thinspace $}

\newcommand{\matr}[1]{\mathbf{#1}} % undergraduate algebra version
\newcommand{\shortcite}{\cite}

\title{NMT-based Cross-lingual Document Embeddings}
\name{Wei Li \textnormal{and} Brian Mak}
\address{Department of Computer Science and Engineering \\
        The Hong Kong University of Science and Technology
\email{\{wliax, mak\}@cse.ust.hk}}

\begin{document}
\maketitle
 
\begin{abstract}
    %Cross-lingual document embeddings allow many natural language processing applications to work readily with multiple languages using a common embedding framework. 
%A cross-lingual NTDV model can be trained with parallel corpora that are unrelated to the task at hand. Then the task can be developed with NTDVs of one language, and tested with NTDVs of another language. 
This paper investigates a cross-lingual document embedding method that improves the current {\em \underline{N}eural machine \underline{T}ranslation framework based \underline{D}ocument \underline{V}ector} (NTDV or simply NV). NV is developed with a self-attention mechanism under the neural machine translation (NMT) framework. In NV, each pair of parallel documents in different languages are projected to the same shared layer in the model. However, the pair of NV embeddings are not guaranteed to be similar. This paper further adds a distance constraint to the training objective function of NV so that the two embeddings of a parallel document are required to be as close as possible. The new method will be called {\em constrained NV} (cNV). In a cross-lingual document classification task, the new cNV performs as well as NV and outperforms other published studies that require forward-pass decoding. Compared with the previous NV, cNV does not need a translator during testing, and so the method is lighter and more flexible.

%More importantly, it paves the way for creating a language-independent embedding for a document written in multiple languages. 

% cNTDV runs fast because it requires only a forward-pass through the encoder of an NMT model to produce document vectors. 
%This paper investigates a document embedding method that is independent of its language.
%%enough so that a translator is needed to produce the cross-lingual embeddings.

%presents an improvement to the previous NTDV method by further 
\end{abstract}

\section{Introduction}
\label{intro}

%Distributed representation of text can help us discover the relationship between words and sentences, categorizing documents and sharing the knowledge between related texts \cite{le2014distributed,mikolov2013efficient,dai2015document,ai2016analysis,NIPS2015_5950}. 
Cross-lingual embedding of  texts from different languages to a unified space will enable comparison and knowledge-sharing between languages
\cite{hermann2014multilingual,ap2014autoencoder,mogadala2016bilingual,berard2016multivec,ferreira2016jointly,gouws2015bilbowa,artetxe2018margin,schwenk2018corpus,sinoara2019knowledge}. With cross-lingual
embedding, data in resource-rich languages can be used to help
understand inputs from resource-scarce languages. Prior works can be categorized into cross-lingual word embedding
\cite{klementiev2012inducing,coulmance2016trans,ap2014autoencoder,berard2016multivec}  and cross-lingual text (document/sentence) embedding
\cite{hermann2014multilingual,pham2015learning,mogadala2016bilingual,berard2016multivec}. 
In this study,  a cross-lingual document/sentence vectorization model for cross-lingual text embedding is developed under a self-attentive neural machine-translation (NMT) framework with a distance constraint (cNTDV, or cNV for short).   
cNV is improved from a recent method called {\em neural machine translation framework based document vector} (NTDV, or NV for short) \cite{li2018fast} with an additional constraint to minimize the distance between their two embedding vectors produced from each parallel document. 
Whereas an NV in a cross-lingual task is formed by combining the two embeddings of a document in its source language and its translated language, these two embeddings produced by our new cNV method are so similar that,
in the production mode, only one is needed (as using both gives only small additional performance gain). Thus, there is an option of using or not using the extra translation step. (NV needs a translator to produce translated text in order to form a joint vector for an application). For better performance (accuracy), the translator could be used; for faster performance, the translator can be removed and the performance is still superior to many traditional forward-propagation methods. cNV is more flexible and lightweight for applications.  Note that the aforementioned translator is trained with the same Europarl corpora as in previous research. Moreover, we also study the effectiveness of distance constraint training by visualizing the manifold projection of various NV and cNV embedding.

The cross-lingual document classification task on RCV1/RCV2 was used to evaluate the effectiveness of the cNV model \cite{lewis2004rcv1,klementiev2012inducing}. If the cross-lingual document embedding method is successful, the classifier (e.g., SVM) trained on 1K English embeddings would be able to classify the 5K German embeddings in the test set as well. Note that the embedding models were pre-trained on the Europarl v7 parallel corpus which is unrelated to this classification task. This design reflects the reality that tasks unrelated parallel data is easy to find while task-related user data for a new language is harder to obtain. The embedding model is not fine-tuned using any testing or training data in the RCV1/RCV2 dataset. This is inherently different from the task-unrelated pre-training and task-specific fine-tuning framework like XLNet \cite{yang2019xlnet}. Albeit the pre-training and fine-tuning framework would normally produce better performance with encoder and classifier optimized together in a complete network, in many NLP applications, it is more convenient to have a readily available pre-trained embedding without fine-tuning so a single uniform embedding can be swiftly plugged into many different applications, such as comparison, clustering or indexing. It would be much easier to compare or index uniformed embedding vectors across tasks than the different encoder outputs fine-tuned in each task.   

There are also NMT based approaches like LASER \cite{schwenk2017learning,artetxe2019massively}, where the cross-lingual embedding can be obtained by using a uniform dictionary, shared encoder, and shared target languages without using a distance constraint. However, one motivation of the cNV and NV model is that we can easily convert the models from a pair of existing NMT models. Shared encoder and uniform dictionary setup are used in multi-lingual translation models which are rare to find. Most existing NMT models only translate between two languages. A pair of NMT models normally would have different encoder/decoder and dictionary indexing, so we have to assume that they are not shareable. Therefore, NV and cNV would use independent encoders and dictionaries in different translation directions. When training the cNV and NV, we only need to copy the encoder and decoder parameters from a pair of NMT models and then adapt the shared layers with a few iterations. Moreover, after pre-training, we can also use the existing NMT models to boost the performance of our model as mentioned in the previous paragraph. Finally, the uniform dictionary in LASER would multiply the input/output layer size by the number of languages the multi-lingual translation model includes. Fully sharing an encoder/decoder across different language families (e.g. Chinese and English) could potentially cause a drop of performance in a multi-lingual model \cite{sachan2018parameter}. A LASER-like framework is more suitable for a large number of languages while NV and cNV are designed to be easily adapted from pre-trained NMT models which are abundantly available. Note that the pre-trained encoder provided in LASER \cite{schwenk2017learning,artetxe2019massively} is a much deeper model and is trained with far more data than the standard Europarl corpus which is used in this study and many previous studies.

%Moreover, compared to XLnet, this setup would also provide flexibility to the choice of classifier and task, a user can change to other types of classifiers (e.g. perceptrons) or tasks without having to fine-tune the embedding model.

%(a) The basic cNTDV model. (b) The production of cNTDV$_{any}$ and NTDV$_{any}$ when the input text is English(left) and  German(right). (c) The production of cNTDV$_{en}$ and NTDV$_{en}$ when the input text is English(left) and  German(right). (d) The production of cNTDV$_{en+de}$, cNTDV$_{con}$, NTDV$_{en+de}$ and NTDV$_{en:de}$ when the input text is English.
%skls

%The encoder has a word embedding layer and a bi-directional gated recurrent unit (GRU) layer; the decoder has a word embedding layer and a conditional GRU ($cGRU_{att}$) layer. 

%\begin{figure*}
%   \begin{center}
%     \begin{tabular}[t]{cc}
%       \subfigure[The basic cNTDV model ]{\resizebox{0.45\textwidth}{!}{\includegraphics{cNTDV.PNG}}} 
%       &
%       \subfigure[The stacked cNTDV$^2$ model]{\resizebox{0.45\textwidth}{!}{\includegraphics{cNTDV2.PNG}}} 
%     \end{tabular}
%   \end{center}

 %   \caption{The cross-lingual self-attentive document/sentence vectorization model.}
 %   \label{fig:cNTDV1_2}
%\end{figure*}

\begin{figure*}
    \vspace{-1.2cm}
   \begin{center}
     \begin{tabular}[t]{cc}
        \vspace{-3mm}
        \hspace*{-4.00em}
        %\scriptsize{
        \subfigure[ \scriptsize{In training, English inputs pass through the left NMT$_{en\rightarrow de}$, while German inputs pass through the right NMT$_{de\rightarrow en}$.}]{\resizebox{0.26\textwidth}{!}{\includegraphics [ height=0.26cm]{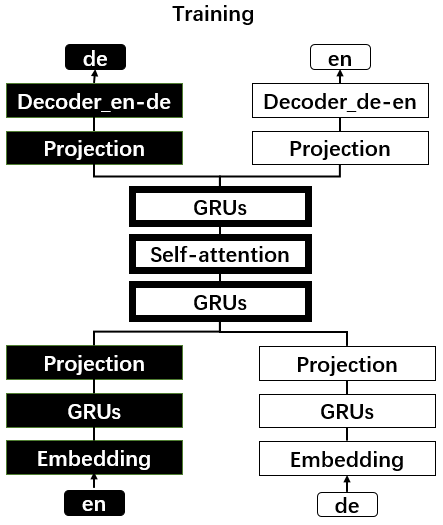}}} 
        %}
        \hspace*{0.20em}
        &
        \subfigure[\scriptsize{Producing \cnvsumconen and  \nvsumconen with English (left) or German (right) input texts. German input is first translated to English to produce $\matr P^{(en)}$. }]{\resizebox{0.26\textwidth}{!}{\includegraphics [ height=0.26cm]{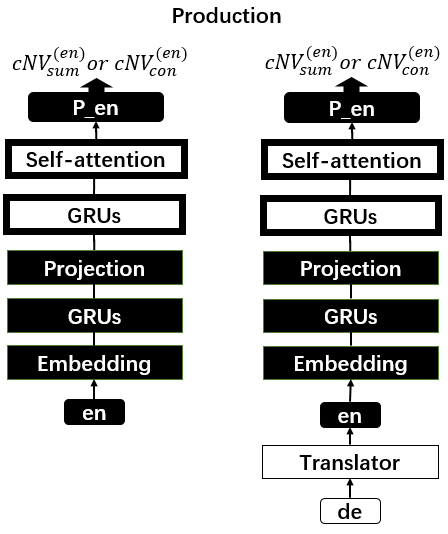}}}
     
      \hspace*{0.20em}
      
       \subfigure[\scriptsize{\cnvsumcony is  \cnvsumconen if the input is English (left). \cnvsumcony is \cnvsumconde if the input is German (right). No translation is needed.}]{\resizebox{0.26\textwidth}{!}{\includegraphics [ height=0.26cm]{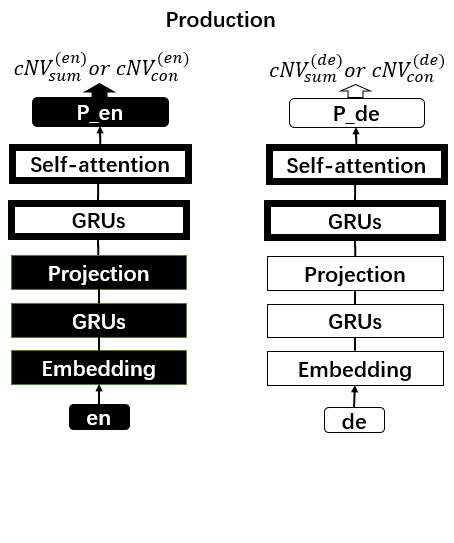}}}
        
      %\subfigure[\scriptsize{Left:
      %cNV$^{(en)}_{sum}$ or cNV$^{(en)}_{con}$ constructed from $\matr P^{(en)}$ with English input. Right: 
      %cNV$^{(de)}_{sum}$ or cNV$^{(de)}_{con}$ constructed from $\matr P^{(de)}$ with German input.}]{\resizebox{0.26\textwidth}{!}{\includegraphics [ height=0.26cm]{cNV_any_pro.PNG}}}
      %\subfigure[\scriptsize{Left: cNV$^{y}_{(con)}\!\!\!\!=\!\!$ cNV$^{en}_{(con)}$ (from $\matr P^{(en)}$) when input is English. Right: cNV$^{y}_{(con)}\!\!\!=\!$ cNV$^{de}_{(con)}$ (from $\matr P^{(de)}$) when input is German.}]{\resizebox{0.26\textwidth}{!}{\includegraphics [ height=0.26cm]{cNTDV_pro_any.PNG}}}
   
       \hspace*{0.20em}
       
       \subfigure[\scriptsize{Producing \nvori, \nvsum , \cnvsum and \cnvcon when input text is English. $\matr P^{(en)}$ and $\matr P^{(de)}$ are combined to obtain the final document vector.} ]{\resizebox{0.26\textwidth}{!}{\includegraphics [ height=0.26cm]{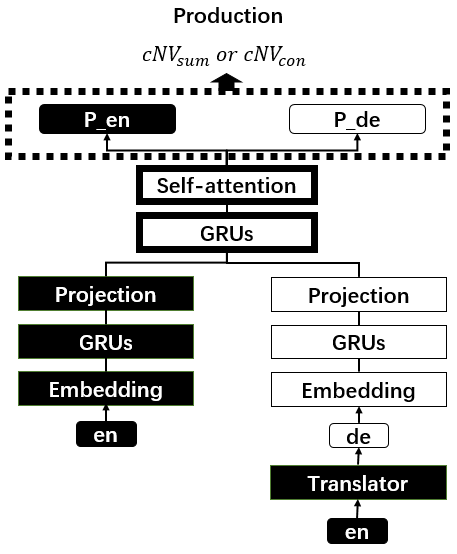}}}

     \end{tabular}
   \end{center}
   \vspace{-0.3cm}
    \caption{The differences between various NV and cNV document vectors in training and production modes.} 
    \vspace{0.4cm}
    \label{fig:cNTDV_proX4}
\end{figure*} 

\section{Model architecture}
\label{model}

\subsection{The attention-based NMT model}
%\vspace*{3mm}

The cNV model is built on the NMT framework, which consists of an encoder-decoder structure with an attention mechanism \cite{sennrich2017nematus}. cNV shares many similarities with a bilingual NMT model. It has two NMT models (NMT$_{a\rightarrow b}$ and NMT$_{b\rightarrow a}$) in opposite directions, where $a\rightarrow b$ indicates the translation from source language $a$ to target language $b$. Shared layers are then inserted between the encoders and decoders. Fig. \ref{fig:cNTDV_proX4}(a) shows the cNV model in the training mode with $a$ being English (en) and $b$ being German (de). NMT$_{a\rightarrow b}$ and NMT$_{b\rightarrow a}$ are first trained independently. Then their parameters are copied into the cNV model and are fixed in subsequent cNV training, in which only the parameters of the shared layers are updated. Given a pair of English and German sentences, the English sentence will be processed through the encoder of NMT$_{en\rightarrow de}$, the shared layers, and its decoder resulting in the translation loss of $l_{mt}^{en\rightarrow de}$. Similarly, the German sentence in the pair will be processed through the encoder of NMT$_{de\rightarrow en}$, the same shared layers, and its decoder resulting in the translation loss of $l_{mt}^{de\rightarrow en}$. 
%Inspired by the transformer architecture in NMT, we also investigate a stacked self-attention structure (named as cSV) which replaces the shared GRU layer by another self-attention layer.

Previous researches show that the information from the encoder helps decide the content of the translated text --- the {\em adequacy} \cite{tu2016context,ding2017visualizing}. The NV model uses a multi-head self-attentive mechanism to summarize the adequacy information, taking advantage of the information extraction ability of the self-attention layer. This self-attention layer focuses only on the overall sequential pattern, in contrast to the self-attention mechanism in the transformer network that puts more weight on the time-sensitive-information \cite{vaswani2017attention}. It summarizes the input sequence to a fixed-length matrix (which becomes the document representation after some post-processing). Let $\vec z_i $ be the hidden state of the shared GRU layer. The context vector $\vec p_m$ produced by the $m$th attention head is given by:
\begin{eqnarray}
g^m_i \!\!\! & = \!\! & f \left(\frac{ ( \matr W_q^m \vec { z_i}  + \vec b_q^m)^T (
    \matr W_k^m \vec {z_i} + \vec b_k^m)} {\sqrt{d_h}} \right)
\label{eq5} \\           
\vec p_m \!\!\! & = \!\!\!\! & \sum_{i=1}^{L_x} g^m_i ( \matr W_v^m \vec { z_i} + \vec b_v^m) \label{eq6}   
\end{eqnarray}
where $f(\cdot)$ is the softmax function;  $i$ is the input word index; $\matr W_q^m$, $\matr W_k^m$, $\matr W_v^m$ and $\vec b_q^m$, $\vec
b_k^m$, $\vec b_v^m$ are model parameters of the $m$th attention head, and the subscripts $q$, $k$, $v$ indicate the role of {\em query}, {\em key} and {\em values} in the attention mechanism \cite{vaswani2017attention}, respectively; $d_h$ is the hidden layer size; $L_x$ is the length of input sentence.

Let there be $r$ attention heads, and
$\matr P^{(a)}$ and $\matr P^{(b)}$ be the context matrix $\matr P$ from NMT$_{a\rightarrow b}$ and NMT$_{b\rightarrow a}$, respectively, in which the $m$th column vector represents the context vector from the $m$th head. Different document/sentence vectors can be extracted by summing or concatenating the column vectors in $\matr P^{(a)}$ and/or $\matr P^{(b)}$ as follows: 
\begin{eqnarray}
\mathrm{cNV}^{(a)}_{sum} \! & = \!  & \sum_{i=1}^{r} \vec p_i^{\ (a)} \label{eq11} \\
%\mathrm{cNV}_b \! & =  \! & \sum_{i=1}^{r} \vec p_i^{\ (b)} \label{eq12} \\
\mathrm{cNV}_{sum}\! & =  \! & \mathrm{cNV}^{(a)}_{sum}+\mathrm{cNV}^{(b)}_{sum} \label{eq13} \\
%\mathrm{cNV}_{:} & =  \! & concat(\mathrm{cNV}_a, \mathrm{cNV}_b) \label{eq14} \\
\mathrm{cNV}^{(a)}_{con} \! & =\! & concat(\vec p_1^{\ (a)},\vec p_2^{\ (a)},\ldots,\vec p_r^{\ (a)}) \label{eq21} \\
%\mathrm{cNV}^{(b)}_{con} \! & =  \! &  concat(\vec p_1^{\ (b)},\vec p_2^{\ (b)},\ldots,\vec p_r^{\ (b)}) \label{eq22} \\
\mathrm{cNV}_{con}\!  & =  \! & \mathrm{cNV}^{(a)}_{con}+\mathrm{cNV}^{(b)}_{con} \label{eq23} \
\end{eqnarray}
where $\vec p_i^{\ (a)}$ and $\vec p_i^{\ (b)}$ are the $i$th column of $\matr P^{(a)}$ and $\matr P^{(b)}$, respectively, and $concat()$ is vector concatenation. Moreover, any operation for language $a$ also applies to language $b$.

 %\begin{figure*}
   %\vspace{-0.8cm}
   %\begin{center}
     %\begin{tabular}[t]{cc}
       %\hspace*{-4.00em}
       %\subfigure[cNV ]{\includegraphics[width=.31\textwidth]{a500.png}} 
       %\hspace*{-3.00em}
       %&
       %\subfigure[NV ]{\includegraphics[width=.31\textwidth]{b500.png}}
       %\hspace*{-3.00em}
     
       %\subfigure[cNV ]{\includegraphics[width=.31\textwidth]{a500bi.png}}
       %\hspace*{-3.00em}
       
       %\subfigure[NV ]{\includegraphics[width=.31\textwidth]{b500bi.png}}
       
       %\subfigure[cNTDV]{\resizebox{0.27\textwidth}{!}{\includegraphics{a500bi.png}}} 
       %\subfigure[NTDV ]{\resizebox{0.27\textwidth}{!}{\includegraphics{b500bi.png}}} 
     %\end{tabular}
   %\end{center}
    %\vspace{-0.6cm}
    %\caption{ The distribution of embeddings from the four document classes and the distribution of English and German embeddings in the test set. }
    %\vspace{0.2cm}
    %\label{fig:dist_pair500}
%\end{figure*}

\subsection{Training with a distance constraint}

Unlike the NV model in \cite{li2018fast}, we would like to have \cnvsumcony
(where the $y$ ($any$) superscript indicates either $en$ or $de$ language, see Fig. \ref{fig:cNTDV_proX4}(c)); The $\matr P^{(en)}$ and $\matr P^{(de)}$ produced by our new cNV model for the same document in either language of a language pair are to be very similar, so that one may use either one or both of them for cross-lingual tasks. To do that, a distance cost is added to the training cost function. The idea is to minimize the distance between related sentence pairs \{$\matr P^{(a)}$, $\matr P^{(b)}$\}  while maximizing the distance between unrelated sentence pairs \{$\matr P^{(a)}$, $\matr P^{(b_j)}$\}, where $\matr P^{(b_j)}$ is the context matrix obtained from a randomly sampled $j$th sentence  \cite{hermann2014multilingual}. The distance cost with one negative sample is:

\begin{equation}
l_{d_j}=\max(0,\alpha +\alpha \frac{k_{d_j}}{2*v_{norm}+1} ) \ ,
\label{eq43}
\end{equation}
where 
\begin{equation}
k_{d_j}=|\matr P^{(a)}-\matr P^{(b)}|^2-|\matr P^{(a)}-\matr P^{(b_j)}|^2 \ ,
\label{eq42}
\end{equation}
$\alpha$ is the margin and $|\cdot |$ is the Frobenius norm of a matrix; $v_{norm}$ is the average Frobenius norm of vectors in the training batch, `+1' is to prevent division by zero . Finally, the total cost of the model is the sum of the distance costs from $N_s$ samplings plus the translation cost: 
\begin{equation}
  l_{com}=\beta \sum_{j=1}^{N_s} {l_{d_j}} + (1-\beta) (l_{mt}^{\ a\rightarrow b}+l_{mt}^{\ b\rightarrow a} ) \ ,
\label{eq44}
\end{equation}
where $l_{mt}^{\ a\rightarrow b}$ and $l_{mt}^{\ b\rightarrow a}$ are the cross-entropy losses of the outputs from NMT$_{a\rightarrow b}$ and NMT$_{b\rightarrow a}$, respectively. $0 \leq \beta \leq 1$ is a weight to balance the translation cost and the distance cost. This novel cost function ensures that the produced document vectors contain ample semantic information and are, at the same time, very similar for parallel documents in a language pair. 

%%%%%
Fig. \ref{fig:cNTDV_proX4} shows the production of various NV/cNV vectors. The main difference between them is that the translator is necessary in the previous NV model but is optional in the cNV model. In comparison with \nvsumconen, \nvsumconde  and \nvsumcon  \footnote{This paper adopts a different set of notations. \nvori  is DV$_{en:de}$ and \nvsum is DV$_{en+de}$ in \cite{li2018fast}. Subscript $\star$ indicates either con, sum, or `:'. \nvori concatenates both \nvsumen and \nvsumde. We stop using `:' in this paper as it always requires both pathways (it does not have `an(y)' option). Moreover, it is difficult to extend \nvori beyond two languages.}, this paper's \cnvsumcony do not use NMT translators. Instead,  \cnvsumcony only use the NMT$_{en\rightarrow de}$ encoder for English inputs and the NMT$_{de\rightarrow en}$ encoder for German inputs. The corresponding \nvsumcony would have a very bad performance since  \nvsumconen and  \nvsumconde from the same document are not guaranteed to be similar. 
%%%%%

%\input{tbl-dim}

%\begin{table}[tbph]
%	\begin{center}
%		\caption{\it The setting of various vectors in the stacked cNTDV$^2$ model.}
%		\begin{tabular}{|c|c|c|c|} \hline
%			Feature & Dimension & $r$ & $d_h$  
%			\\ \hline \hline
%			% $d_h$     & 1024 \\ \hline
%%			cNTDV$_{en+de}^2$  & 500 & 8 & 500\\ \hline
%			cNTDV$_{any}^2$  & 500 & 8 & 500\\ \hline
%		\end{tabular}
%		\label{tbl:dimension_stacked}
%	\end{center}     
%\end{table}   

\section{Experiments}
\label{Experiment}

\begin{table}[tbph]
	\begin{center}
		\vspace{-2mm}
		\caption{\it The setting of various vectors in the cNV model.}
	    \vspace{-2mm}
		\begin{tabular}{|c|c|c|c|} \hline
			Feature & Dimension & $r$ & $d_h$  
			\\ \hline \hline
			
			  \cnvsum & 1024 & 4  & 1024\\ \hline
			 \cnvsumy & 1024 & 4 & 1024\\ \hline
			 \cnvcon & 4096  & 4 & 1024\\ \hline
			 \cnvcony & 4096 & 4 & 1024\\ \hline
		\end{tabular}
		
		\label{tbl:dimension}
		\vspace{-2 mm}
	\end{center}     
\end{table}

%\subsection{Training and text processing}
%, which consists of 1.9M parallel sentences with 49.7M English tokens and 52M German tokens
Our embedding model was trained on the Europarl v7 parallel corpus \cite{koehn2005europarl}. For training the attention-based NMT models and cNV, we adopt the same settings as in \cite{li2018fast} with the following additional settings due to the distance constraint: $d_h=1024$, $r=4$, $N_s = 20$, $\beta = 0.5$, $\alpha = 2\sqrt{d_h}$. In training the attention-based NMT models, we use the default setting with a mini-batch size of 80, a vocabulary size of 85,000, a maximum sentence length of 50, word embedding size of 500, and hidden layer size of 1024. The models are trained with the Adam optimizer using the cross-entropy
loss. Model parameters of cNV are copied from this pair of pre-trained NMT models. The three newly added shared layers are further trained for 5 epochs with the original NMT
parameters fixed. Note that if third-party pre-trained NMT models are available, we only need to fine-tune the three shared layers with 5 epochs. Here we train the NMTs from scratch for fair comparisons. 

The cross-lingual document classification task on RCV1/RCV2 was used to evaluate the effectiveness of the cNV model\cite{lewis2004rcv1,klementiev2012inducing}. In this task, about 1K English documents of four categories are converted to word/document embeddings and are given to classify the category of 5K German documents in the test set, and vice versa \footnote{11980 in total}. If an SVM classifier (we use the same setting as in \cite{li2018fast}) is trained on embeddings from 1K English data and can accurately classify embeddings from 5K German data, then it indirectly proves that English and German texts are projected into the same vector space. Note that the 1K training set here is used to train the SVM classifier, whereas the cNV model is pre-trained on task-independent  Europarl corpus. 

Six-fold cross-validation was used to further examine the statistical significance of the results. The 6 train sets were constructed from the original train set plus 5 train sets sampled from the 5K test data of its parallel task (English from the de$\rightarrow$en test set and German from en$\rightarrow$de test set). 
The single sample t-test was conducted and differences between any two results are significant when $p<0.01$.

%The mean and standard deviation of the six tests is presented as superscript and subscript in Table \ref{tbl:results.accuracy}. 
 %For direct proof, we also visualize the cNV embedding of different languages using their t-SNE projections in Fig. \ref{fig:dist_pair500} and \ref{fig:dist_pair50}.  

\label{Result}

 \begin{figure*}
   \vspace{-1.2cm}
   \begin{center}
     \begin{tabular}[t]{cc}
       \hspace*{-5.00em}
       \subfigure[\cnvsumen ; \cnvsumde ]{\includegraphics[width=.29\textwidth]{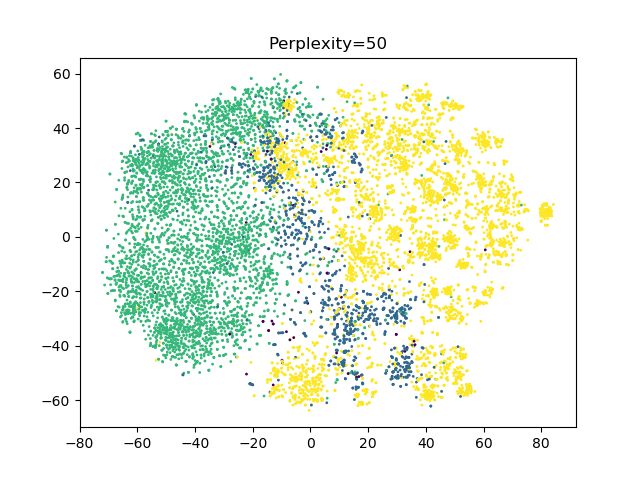}} 
       \hspace*{-2.9em}
       &
       \subfigure[\nvsumen; \nvsumde ]{\includegraphics[width=.29\textwidth]{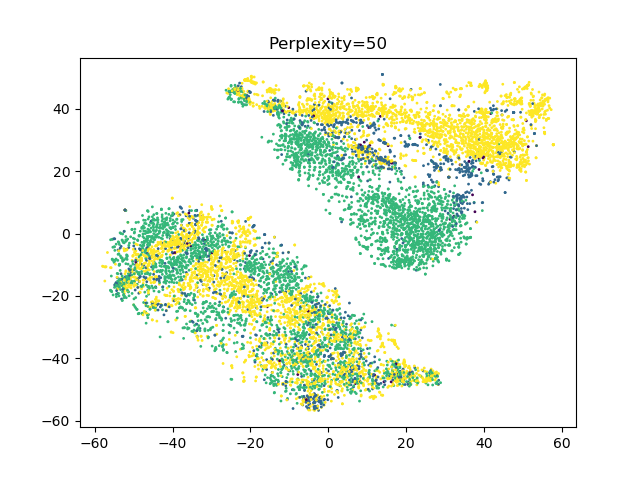}}
       \hspace*{-2.em}
     
       \subfigure[\cnvsumen ; \cnvsumde ]{\includegraphics[width=.29\textwidth]{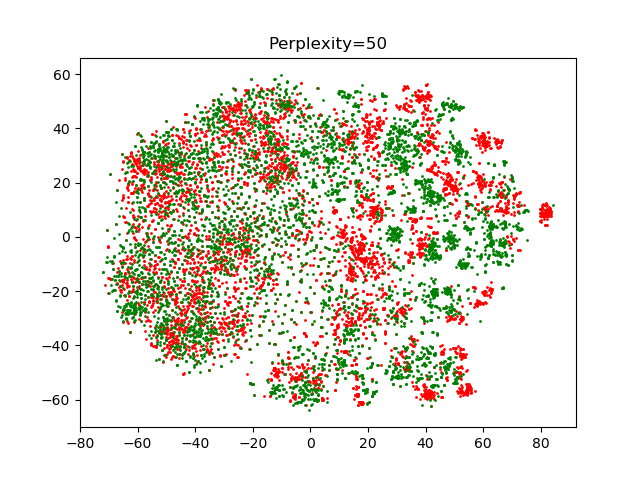}}
       \hspace*{-2em}
       
       \subfigure[\nvsumen ; \nvsumde ]{\includegraphics[width=.29\textwidth]{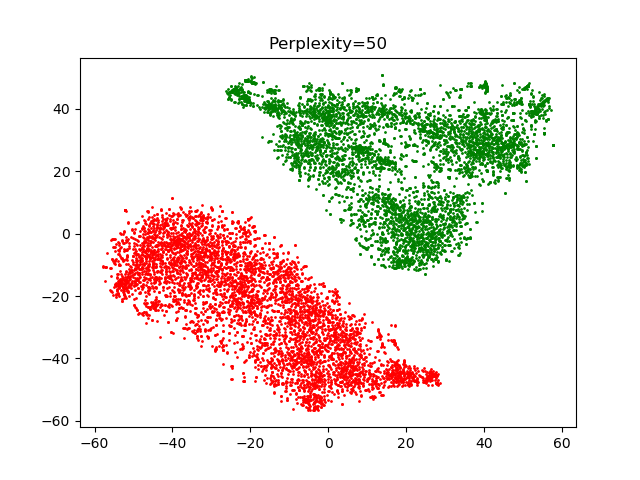}}
       
       %\subfigure[cNTDV]{\resizebox{0.27\textwidth}{!}{\includegraphics{a500bi.png}}} 
       %\subfigure[NTDV ]{\resizebox{0.27\textwidth}{!}{\includegraphics{b500bi.png}}} 
     \end{tabular}
   \end{center}
    \vspace{-0.6cm}
    \caption{The distribution of embeddings from the four document classes (subfigure a, b) and the distribution of English and German embeddings (subfigure c, d) in the test set. }
    \vspace{0.2cm}
    \label{fig:dist_pair500}
\end{figure*}

\begin{figure*}[tbhp]
\vspace{-3mm}
   \begin{center}
     \begin{tabular}[t]{cc}
       \subfigure[%With distance constraint training 
       \cnvsumen ; \cnvsumde]{\resizebox{0.40\textwidth}{!}{\includegraphics{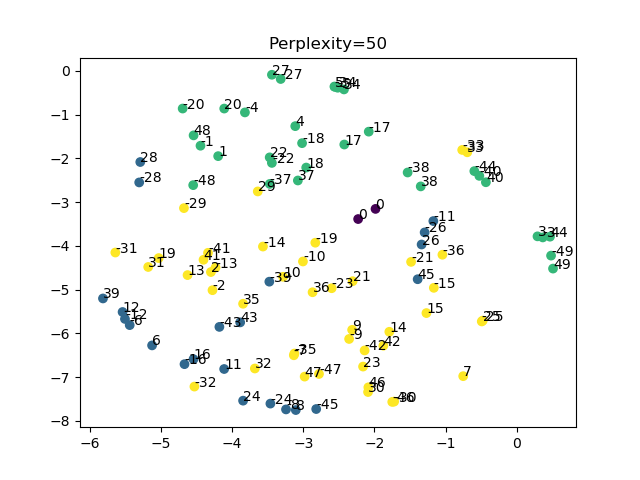}}} &
       
       \subfigure[%Without  distance constraint training 
       \nvsumen ; \nvsumde]{\resizebox{0.40\textwidth}{!}{\includegraphics{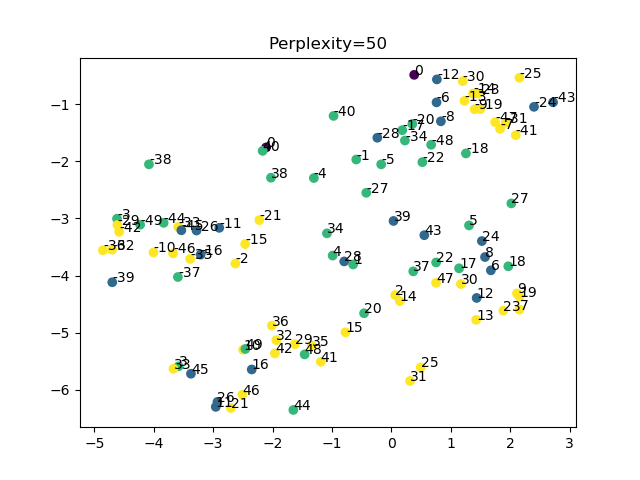}}} 
     \end{tabular}
   \end{center}
    \vspace{-0.6cm}
    \caption{English (+ve) and German (-ve) embedding pairs of 50 randomly sampled documents.}
    %\vspace{0.2cm}
    \vspace{0.5cm}
    \label{fig:dist_pair50}
\end{figure*}

\section{Results}

\begin{table}[tbph]
%\vspace*{-2mm}
\hspace*{-2.50em}
  \begin{threeparttable}
  %\begin{center}
    \caption{\it Classification accuracy (\%) on CLDC   }
    \vspace*{-2mm}
    %\vspace{2mm}
     %\resizebox{\textwidth}{!}{%
      
     \begin{tabular}{| c | c | c | c| c| c|} \hline
      Method &  en$\rightarrow$de &  de$\rightarrow$en & Method & en$\rightarrow$de &  de$\rightarrow$en \\ 
      \hline \hline
      % I-Matrix & 77.6 &  71.1 & - \\ \hline
      % NMT\_base   &  X  & X & parallel \\ \hline
      
      %& 
      % NV$_{+}$  \shortcite{li2018fast} @ & 93.5   & 82.3 

      MT$_{bs}$ \shortcite{klementiev2012inducing}& 68.1 &  67.4 &  \footnotesize \nvsum   \shortcite{li2018fast} @ & 93.5   & 82.3  \\
      NMT$_{bs}$  & 92.9 &  70.8 &  \nvori \shortcite{li2018fast}  @ & \bf 94.4  & 82.7\\
      
       \footnotesize$para\_doc$ \shortcite{pham2015learning}*& 92.7 &\bf 91.5 &   \cnvsumen @   & 93.1  & 82.8 \\
      BAE \shortcite{ap2014autoencoder}  & 91.8 & 74.2 &  \cnvsumde @  & 93.2  & 81.4 \\
      
      ADD \shortcite{hermann2014multilingual}    & 86.4 & 74.7  & \cnvsum @ & 94.1  & 83.0  \\
      BI \shortcite{hermann2014multilingual}    & 86.1 &  79.0 & \cnvcon @ &  \bf 94.3  & \bf 83.3  \\
      
      % cSV$_{+}$ @ & 93.0 & 81.4 
     
      BRAVE \shortcite{mogadala2016bilingual}    & 89.7& 80.1  & \nvsumy   &  35.6 & 41.9 \\
      MultiVec \shortcite{berard2016multivec}    & 88.2 & 79.1    & \cnvsumy   &  88.9 & 79.1 \\

      Unsup \shortcite{luong2015bilingual}  & 90.7 & 80.0 & 
      \cnvcony    &  \bf 89.8 & \bf 81.0   \\
     
      %NV$_{:}$  \shortcite{li2018fast}  @ & \bf 94.4  & 82.7 &   \\ 
     
      % & cSV$^{y}$    & \bf 91.1 & 80.9 
     
      \hline
       %cNV$_{con}^{y}$    &  89.8$^{\;90.2}_{\;\pm0.65}$ & \bf81.0$^{\;81.1}_{\;\pm0.23}$   \\

      %cNV$_{con}$ @ & \small{ \bf 94.3$^{\;94.0}_{\;\pm0.34}$}  & \bf83.3$^{\;83.6}_{\;\pm0.60}$   &

       \hline

      \end{tabular}
      %}
      
  \begin{tablenotes}
  \small
  %\footnotesize
  \item[]\footnotesize { \cnvsumconen (Fig. \ref{fig:cNTDV_proX4}(b)) only goes through the en$\rightarrow$de path (if the input is German, it is first translated into English by NMT); similar for \cnvsumconde. \cnvsumcony (Fig. \ref{fig:cNTDV_proX4}(c)) only goes through the en$\rightarrow$de path if the input is English, and only the de$\rightarrow$en path if the input is German; no translation is needed. On the other hand, \cnvsumcon (Fig. \ref{fig:cNTDV_proX4}(d)) is the sum of \cnvsumconen  and \cnvsumconde , so the translation is needed in computing one of them. NMT is trained with the same Europarl corpus. BAE adds monolingual test data in the training, whereas others do not. 
  
 \vspace*{1.00em}
  
  %$sum$ or $con$ subscript: sum or concatenate the context vectors over all attention heads. 
  %On the 
 % other hand, cNV$_{\star}$ is the sum of cNV$^{(en)}_{\star}$ and cNV$^{(de)}_{\star}$, and so translation is needed in computing one of them.
  %NMT is trained with the Europarl corpus.  BAE adds monolingual test data in the training, whereas others do not. 
  }
  \end{tablenotes}
  \label{tbl:results.accuracy}
  %\end{center}
  \end{threeparttable}
\end{table}
\vspace*{0.00em}

% `*' uses iterative BP optimization during the production of document vectors. BAE add monolingual test data in the training. `@' uses a forward pass of NMTs (trained with Europarl) to enhance performance (i.e. `NMT-assisted-forward' methods). Other results are trained with only Europarl and only employ one-time forward pass of their respective model to produce

Table \ref{tbl:results.accuracy} presents the performance of various embedding methods on the cross-lingual document classification task. They can be categorized into 3 types of methods:  `*' methods use iterative back-propagation(BP) optimization during the production of document vectors; `@' methods use a forward pass of NMTs (trained with Europarl) to enhance performance (i.e., `NMT-assisted-forward' methods); other results are trained with Europarl and only employ one-time forward pass of their respective model (i.e., forward method). MT$_{bs}$ is the machine translation baseline from \cite{klementiev2012inducing}. NMT$_{bs}$ is the performance of the NMT model on which the cNV model was built. In NMT$_{bs}$, the classifier was trained with %TF/IDF 
the term frequency/inverse document frequency 
of the most frequent 50000-word features. Among all the results,  \cnvcon is significantly better than $para\_doc$ ($p=0.000215$) in en$\rightarrow$de. For de$\rightarrow$en, our \cnvcon gives the second-best result. It is also more convenient to use than methods (e.g., $para\_{doc}$) that require BP in production mode. As BP training is done iteratively and involving derivative calculations, it is not fit for situations that require fast online decoding. Moreover, the vectors produced by BP training will not be consistent in different runs or different systems (e.g., with different training epochs, random initialization, different batch sizes). Thus the forward pass method has a unique advantage in many applications.

 The cNV model use either the `NMT-assisted-forward' method or the `forward' method. This flexibility is made possible because the embeddings produced by the cNV model from the same document in a language pair are very similar, and one may use either one of them or both for cross-lingual document classification. \cnvcon shares similar good results with \nvori ($p=0.022064$ in en$\rightarrow$de, $p=0.011092$ in de$\rightarrow$en). Among the `forward' methods, \cnvcony  is not significantly worse than the `Unsup' method ($p=0.143811$) in en$\rightarrow$de. \cnvcony is significantly better than the `BRAVE' method ($p=0.000169$) in de$\rightarrow$en. Therefore, \cnvcon and \cnvcony are among the best methods in `NMT-assisted-forward' and `forward' categories, respectively. The \cnvcon result shows that using the NMT and producing document vectors together with cNV is better than using the NMT alone (NMT$_{bs}$). When we want to have the best performance, we can use the \cnvcon model. On the other hand, when we want a lighter model and faster decoding, we can use \cnvcony without the NMT. Note that we do not need to change the model structure, nor do we need to retrain the cNV model when shifting from \cnvcon to \cnvcony; the only difference between these two modes is whether we plug in the translated result from NMT into the other encoder. On the other hand, NV model must use the embedding from the same language (or the combination of two embeddings) in training and testing (please see Fig. \ref{fig:cNTDV_proX4}) and all good NV results belong to the `NMT-assisted-forward' methods. We further verify the poor performance of \nvsumcony as shown in Table 2.

 Without the distance constraint, the English embedding and its corresponding German embedding of the same document is not guaranteed to lie in the same vector space and cannot be compared. To further illustrate this, we visualize the document embeddings of different languages using their t-SNE projections in Fig. \ref{fig:dist_pair500} and \ref{fig:dist_pair50}.
Fig. \ref{fig:dist_pair500} shows the NV/cNV embeddings of German documents in the ${en\rightarrow de}$ test set. We also translate the same documents into a mirrored English set with the NMT to show the distribution of the same documents in a different language. 
In  Fig. \ref{fig:dist_pair500} (a) and (b), the 4 types of documents are plotted in 4 different colors (red: Corporate/Industrial, blue: Economics, green: Government/Social, yellow: Market). The document embeddings are clearly divided into 4 clusters, but only when they are trained with distance constraint then their German embeddings and English embeddings (from the translation of the German texts) are indistinguishable if they belong to the same type. On the other hand, in Fig. \ref{fig:dist_pair500} (c) and (d), English embeddings are shown in red while their German counterparts of the same documents are shown in green regardless of their document types.
It is clear that without the distance constraint training, the NV document embeddings are segregated by the languages into two major clusters.
%, one for the English embeddings from the translated English inputs and NMT$_{en\rightarrow de}$\ encoder, and the other for the German embeddings from the German inputs and NMT$_{de\rightarrow en}$ encoder.
These figures show that simply sharing layers in the NV model without distance constraint training is not sufficient to produce similar embeddings from two languages and they cannot be compared.

Fig. \ref{fig:dist_pair50} shows the t-SNE projections of 50 English-German document embedding pairs. The positive and negative indices refer to the English and German embeddings produced from the same indexed document in the German test set, respectively. For example, `-2' is the German embedding of the 2nd document in the test set, whereas `2' is the English embedding of the same 2nd German document after it is translated to English. It shows that cNV model, trained with the distance constraint, effectively pairs the German and English embeddings from two parts of the model close together, regardless of their different encoders and language inputs. On the other hand, the English and German NV embeddings of the same document are far apart. This also explains why the previous NV model must use an NMT as a bridge to achieve good classification performance, and why in the current cNV model the NMT is not necessary.
\section{Conclusion}
\label{conclude}
Compared with the previous NV model where the embeddings are segregated across different languages, cNV embedding pairs are very close to each other. This opens future research directions in the fast conversion of the existing NMT models to cross-lingual text embedding models. 
%In this paper, we present the cNTDV embedding model which makes several improvements from the previous NTDV embedding model, such as a stacked self-attention structure and adding a distance constraint to the training loss function.
%This property of cNTDV embeddings provides much more flexibility for the embedding in practical usage. 

%Comparing to NDTV model,  
%whose performance is better than similar ``forward propagation models'' in other studies. On the other hand, cNTDV$_{con}$ is the best performing model, with an NMT translator integrated into the framework to achieve an even better performance. Compared with the previous NTDV, cNTDV has the flexibility to either use the NMT translator to enhance performance or not to use it to save computationational cost and memory consumption.

%And in our experience, albeit the model producing $DV_{con}$ is more complicated than the NMT\_base model itself, the computation time from the additional encoder path is negligible.  
%
% As it only uses forward propagation in the production phase, the cross-lingual document embedding cNTDV$^2_{any}$ is a lighter and faster method in the production mode when compared with methods that require iterative back-propagation training (e.g., $para\_doc$).  Among other forward-propagation methods, cNTDV is one of the best performing model and is on par with the previous NDTV. In addition, cNTDV has the flexibility to either use an NMT to enhance performance or not to use it to save computational cost and memory consumption. 
%\input{acknowledge.tex}

\bibliographystyle{IEEEtran}
\bibliography{abbrev,mybib}
\end{document}